\PassOptionsToPackage{table,xcdraw}{xcolor} 
\documentclass{bmvc2k}

\title{Beyond Gloss: A Hand-Centric Framework for Gloss-Free Sign Language Translation}

\addauthor{Sobhan Asasi}{s.asasi@surrey.ac.uk}{1}
\addauthor{Mohamed Ilyas Lakhal}{m.lakhal@surrey.ac.uk}{1}
\addauthor{Ozge Mercanoglu Sincan}{o.mercanoglusincan@surrey.ac.uk}{1}
\addauthor{Richard Bowden}{r.bowden@surrey.ac.uk}{1}

\addinstitution{
 Center for Vision, Speech and Signal Processing (CVSSP)\\
 University of Surrey\\
 Guildford, UK
}

\runninghead{S. Asasi, M. Lakhal, O. Sincan, R. Bowden}{Beyond Gloss}

\usepackage{amsmath}  
\usepackage{amssymb}  
\usepackage{wasysym}
\usepackage{graphicx} 
\usepackage{multirow}
\usepackage{arydshln}
\usepackage{colortbl}
\usepackage{array}
\usepackage{makecell} 
\usepackage{xr}
\usepackage{algorithm}
\usepackage{algorithmic}

\begin{document}

\maketitle

\begin{abstract}
Sign Language Translation (SLT) is a challenging task that requires bridging the modality gap between visual and linguistic information while capturing subtle variations in hand shapes and movements. To address these challenges, we introduce \textbf{BeyondGloss}, a novel gloss-free SLT framework that leverages the spatio-temporal reasoning capabilities of Video Large Language Models (VideoLLMs). Since existing VideoLLMs struggle to model long videos in detail, we propose a novel approach to generate fine-grained, temporally-aware textual descriptions of hand motion. A contrastive alignment module aligns these descriptions with video features during pre-training, encouraging the model to focus on hand-centric temporal dynamics and distinguish signs more effectively. To further enrich hand-specific representations, we distill fine-grained features from Hand Mesh Recovery (HaMeR). Additionally, we apply a contrastive loss between sign video representations and target language embeddings to reduce the modality gap in pre-training.  \textbf{BeyondGloss} achieves state-of-the-art performance on the Phoenix14T and CSL-Daily benchmarks, demonstrating the effectiveness of the proposed framework. Our code is available at \url{https://github.com/elsobhano/BeyondGloss}.
\end{abstract}

\section{Introduction}
\label{sec:intro}
Sign languages use hand and body movement alongside facial expressions, mouth gestures and the complex use of space for communication~\cite{p44, p47}. Sign language translation (SLT) aims to convert these visual elements into spoken sentences, enabling communication with non-signers. However, due to the inherent complexities of natural signing and the fundamental difference to spoken languages, SLT faces challenges in accurately recognizing the components of sign and bridging the gap between the visual and textual representations~\cite{p16, csgcr}.
SLT approaches typically fall into two categories: \mbox{gloss-based} and \mbox{gloss-free}. Gloss-based methods~\cite{mmtlb,sltunet,ts-slt} rely on intermediate representations, where sign videos are first translated into a sequence of glosses, written representations of signs using corresponding words in a spoken language, before generating spoken sentences. In contrast, gloss-free methods directly map videos to text without an explicit gloss representation~\cite{p6,p53, fla-llm}. While gloss-based approaches benefit from structured linguistic information, gloss-free methods bypass the need for gloss annotations, making them more flexible but often more challenging. The high cost and effort required for data collection and meticulous gloss annotation have led to increased interest in gloss-free SLT~\cite{sign2gpt,signllms,liang2024llavasltvisuallanguagetuning}. 
\begin{figure*}[]
\begin{center}
\includegraphics[width=\linewidth]{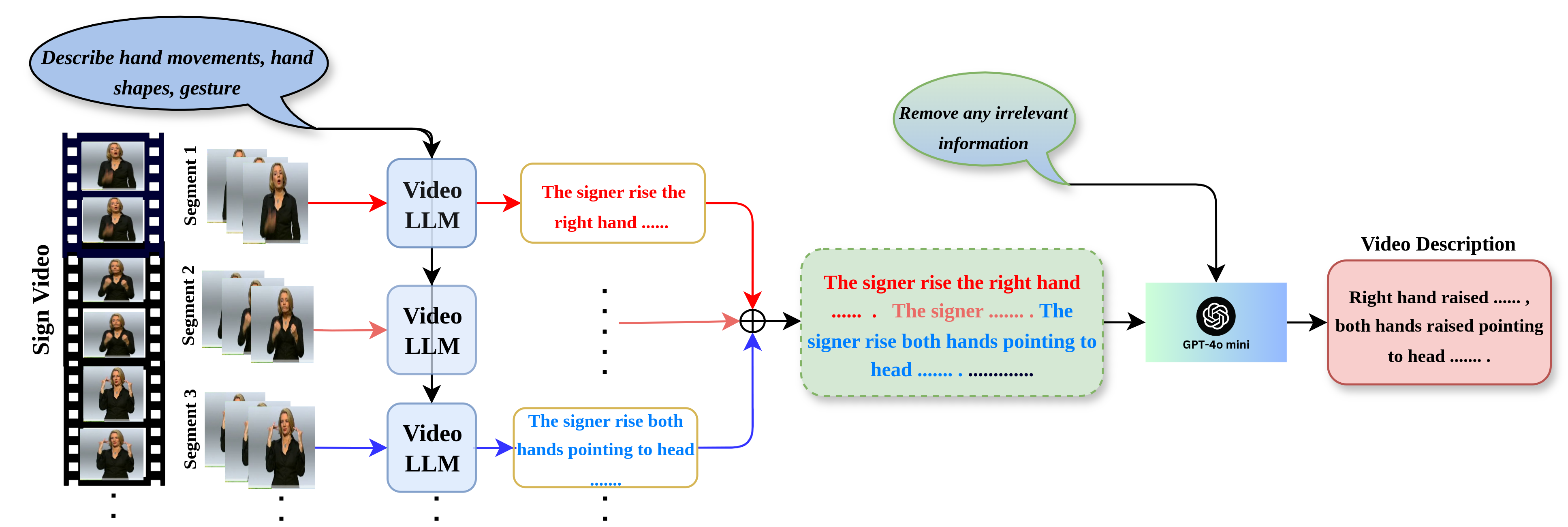}
\end{center}
   \caption{\small \textbf{Proposed Sign Video Descriptor}. The input sign video is first divided into segments that are individually described by a VideoLLM in terms of hand movements, shapes, and trajectories. These segment-level descriptions are then merged and refined by a language model to remove redundancy and preserve temporal order, resulting in a coherent final description that highlights the essential aspects of hand motion across the video.}
\label{fig:overview}
\vspace{-5mm}
\end{figure*}
The increasing use of Large Language Models (LLMs) has had a significant impact on this area~\cite{openai2024gpt4technicalreport}. As a result, the latest gloss-free models utilise the remarkable language prediction capabilities of LLMs by converting sign language videos into text-like representations that LLMs can process effectively~\cite{fla-llm,sign2gpt, gfslt-vlp}.
When LLMs are applied directly to visual features, they become mere transformers with pre-trained weights due to the large shift between the textual representations on which they were trained and the new visual inputs.
To bridge this gap, recent studies have introduced a pre-training phase before the SLT task, which aims to align the visual and textual representations~\cite{sign2gpt, gfslt-vlp, asasi2025hierarchicalfeaturealignmentglossfree}.
Given the central role of hand shape and motion in conveying meaning in sign language, our framework explores the benefit of emphasizing hand-centric features. We achieve this by applying feature distillation from Hand Mesh Recovery (HaMeR)~\cite{pavlakos2024reconstructing}, which guides the model to focus more on hand poses and orientations in individual video frames. We then align the video features with a textual description of the entire video, generated using VideoLLMs, which we employ in the context of SLT. These descriptions provide high-level semantic guidance that is more directly aligned with the visual input than conventional spoken language targets. Finally, to enhance performance in the downstream SLT task, we adopt an encoder-decoder architecture based on language models. To further improve the alignment between the encoder’s output embeddings and the spoken language targets, we incorporate a contrastive learning strategy inspired by the CLIP framework~\cite{clip}, projecting both video and text features into a shared embedding space. 

Our primary contributions can be summarized as: (\textit{i}) Our proposed framework, \textbf{BeyondGloss}, achieves state-of-the-art results in gloss-free sign language translation on two widely used benchmark datasets.  
(\textit{ii}) We introduce a novel method for generating hand-centric textual descriptions of sign language videos using VideoLLMs, offering a new form of semantic supervision aligned with visual content. (\textit{iii}) We are the first to leverage HaMeR’s transformer-based hand representations in the SLT task. Through feature distillation, we guide the visual encoder to focus on hand pose and orientation, demonstrating the positive impact of this strategy on translation quality.

The rest of this paper is organised as follows: In Section \ref{sec:litrature},
we review the literature in sign language translation. In Section \ref{sec:method}, we outline the proposed
\textbf{BeyondGloss} approach. Section \ref{sec:results} presents quantitative and
qualitative evaluation, whilst Section \ref{sec:conclude} concludes.

\section{Related Works}
\label{sec:litrature}
\textbf{Gloss-free Sign Language Translation.}
Gloss-based SLT methods achieve strong performance by using annotated glosses as intermediate supervision~\cite{sltunet, ts-slt}. However, glosses are labor-intensive to create and not available for all languages. Gloss-free SLT avoids this by directly translating sign videos into spoken sentences using only video-text pairs. While more scalable, it faces a larger modality gap and typically performs worse than gloss-based methods.
To address this gap,
CSGCR~\cite{csgcr} predicts possible language words, generates
sentences based on these predictions, and selects the most appropriate sentence using cosine similarity with sign language images. Recently, GFSLT-VLP~\cite{gfslt-vlp} introduced a pre-training method that aligns sign language images with
spoken sentences, converting them into text-like features familiar to LLMs.
Leveraging this advancement, subsequent gloss-free models focus on generating more comprehensible features
for LLMs. FLa-LLM~\cite{fla-llm} introduces a two-step approach: initially training on visual information from sign language
images using a lightweight translation model and fine-tuning the LLM for SLT. Sign2GPT~\cite{sign2gpt} pre-trains a sign
encoder by aligning visual features with prototypes under the supervision of pseudo-glosses, words from spoken sentences via Parts-Of-Speech (POS) tagging, and then utilizes it for SLT. SignLLM~\cite{signllms} designs a vector-quantization
technique to convert sign videos into discrete sign tokens. Lastly, LLaVA-SLT~\cite{liang2024llavasltvisuallanguagetuning} integrates linguistic pretraining, visual contrastive learning, and visual-language tuning within a Large Multimodal Model to effectively bridge the gap between sign language videos and textual translation.\\
\textbf{Video Large Language Models.}
Early VideoLLMs like Video-LLaMA~\cite{videollama} and Video-ChatGPT~\cite{videochatgpt} extended large language models to the video domain by incorporating temporal and multimodal context. Initially limited to short clips and basic temporal understanding, recent models have advanced by combining large-scale video-language data with autoregressive decoding~\cite{vaswani2017attention, graves2013generating}. These systems integrate spatiotemporal encoders with LLMs (\textit{e.g.} LLaMA, GPT) to handle longer sequences and support open-ended video understanding, including tasks like action recognition and captioning.~\cite{llama, NEURIPS2020_1457c0d6}.
Although sign language falls under video understanding, it poses unique challenges due to its reliance on fine-grained, frame-by-frame details, especially in hand shapes, movements, and facial expressions. Current VideoLLMs~\cite{videollama, videochatgpt, li2023videochat, lin2024video} often lose this crucial information due to frame sampling or summarization, and processing all frames at full resolution is computationally costly \cite{weng2024longvlm}. To address this, we propose an approach that leverages VideoLLMs for generating detailed hand-centric descriptions suited for sign language understanding.\\
\textbf{Vision-Language Pre-training.} Existing visual-language pre-training (VLP) models fall into two types: single-stream models, which fuse visual and textual inputs in one encoder using self-attention~\cite{kim2021vilt, li2019visualbert, kamath2021mdetr}, and dual-stream models, which use separate encoders for each modality and align them via contrastive learning or cross-attention~\cite{clip, jia2021scaling, alayrac2022flamingo}. These pre-trained models have demonstrated remarkable performance in downstream tasks~\cite{clip, jia2021scaling}. GFSLT-VLP~\cite{gfslt-vlp} is the first study to utilize CLIP~\cite{clip},
a pioneering approach in VLP, to tackle the modality gap in gloss-free SLT. Like CLIP, GFSLT-VLP adopts a dual-stream structure, pre-training the alignment between sign videos and spoken sentences before performing the translation. \\ \\
\textbf{Hand Reconstruction Modeling.} The typical input to SLT is either RGB video~\cite{nslt, liang2024llavasltvisuallanguagetuning} or skeletal based approach~\cite{gan2021skeleton, jiang2021skeleton}. But recent approaches such as HaMeR~\cite{pavlakos2024reconstructing}, provide potential alternatives that SLT can be built on. Hand Mesh Recovery (HaMeR) is a fully transformer-based framework for 3D hand reconstruction from monocular images, offering enhanced robustness and accuracy by scaling both model capacity and training data~\cite{pavlakos2024reconstructing}. It utilises a large Vision Transformer and integrates multiple datasets with 2D and 3D hand annotations, achieving state-of-the-art results on standard 3D hand pose benchmarks. Given the central role of hand shapes and movements in sign language communication, HaMeR has recently been adopted for sign language understanding tasks. Notably, Hands-On~\cite{he2025hands} was the first to apply HaMeR for sign language segmentation. Inspired by this, our approach similarly leverages HaMeR to extract detailed hand features from video frames, aiding in more accurate sign interpretation.
\vspace{-3mm}
\section{Methodology}
\label{sec:method}
This section presents our proposed \textbf{BeyondGloss} framework. We first describe how high-quality hand features are extracted via distillation in Sec.~\ref{method:distil}, followed by a detailed explanation of our novel Sign Video Descriptor for generating hand motion descriptions in Sec.~\ref{method:descriptor}. Finally, Sec.~\ref{method:arch} outlines the overall architecture for SLT.
\subsection{Distillation}
\label{method:distil}
Distillation~\cite{distil1,distil2,distil3} is a learning technique where a smaller model (student) is trained to replicate the behavior of a larger, well-trained model (teacher). In feature distillation~\cite{distil3, hinton2015distilling}, instead of only imitating final predictions, the student learns to match the intermediate feature representations of the teacher. This helps the student model learn richer internal patterns, improving performance and generalisation with fewer parameters.
In our framework, we apply feature distillation to emphasize hand modeling in sign language videos. Specifically, we align the visual encoder’s features with those from the HaMeR transformer head to enhance sensitivity to hand shapes and orientations. To ensure that broader spatial information is not lost, we combine the hand-focused features with the original ones, maintaining a balance between detailed hand information and overall scene context.

\vspace{-0.4cm}
\subsection{Sign Language Video Description}
\label{method:descriptor}
To encourage our framework to focus on hands and their movements, key components of sign language, we aim to align sign video representations with hand-centric descriptions extracted from the videos. However, due to the complexity of hand motions and the inability of off-the-shelf VideoLLMs to capture fine-grained temporal details (as well as memory constraints)~\cite{weng2024longvlm}, we cannot directly use VideoLLMs by simply feeding them full sign videos. To address these challenges, we propose a novel sign video descriptor that leverages the strengths of both VideoLLMs and LLMs. An example of a generated description is provided in Fig~\ref{fig:overview}. 
As shown in the figure, we begin by segmenting each video into non-overlapping clips of 16 frames. This length is chosen as a balance between temporal resolution and semantic content: segments shorter than 16 frames may lack sufficient motion or variation to be meaningfully described, while longer segments often lead VideoLLMs to produce less accurate or overly generalised descriptions. For each segment, we prompt the VideoLLM to describe hand movements, shapes, and trajectories. After obtaining a separate description for each segment, we merge them into a single sequence and pass them to an LLM that refines the content by removing redundant or irrelevant information. This final description preserves the temporal order and captures the most important aspects of hand motion across the entire video. See Appendix for more details and examples.
\vspace{-0.4cm}
\subsection{Architecture}
\label{method:arch}
\begin{figure*}[!t]
\begin{center}
\includegraphics[width=\linewidth]{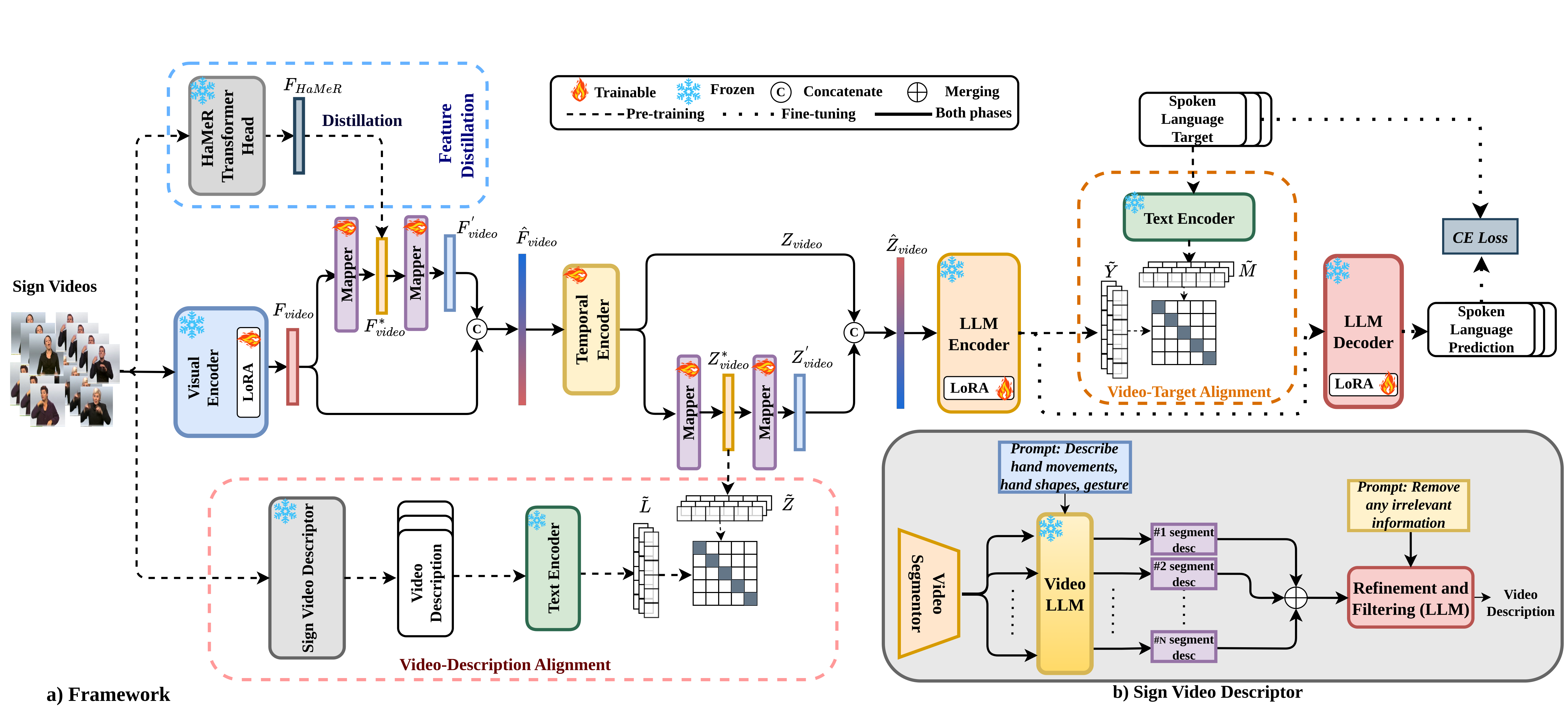}
\end{center}
  \caption{\small \textbf{Overview of BeyondGloss}. a) The framework has two phases: pre-training (dashed+solid lines) and fine-tuning (dotted+solid lines). In pre-training, HaMeR features are distilled, mapped to visual encoder outputs, and fused. After processing by a temporal encoder, the features are mapped to video-description features and fused again. Trainable mappers align feature spaces, removing the need for distillation and alignment modules during fine-tuning. Final features are fed into the LLM encoder and aligned with spoken language, while in fine-tuning, the LLM encoder output is fed directly to the LLM decoder. b) The overview of the proposed Sign Video Descriptor, which segments sign videos, generates descriptions using a VideoLLM, merges them into a sequence, and refines the result through an additional LLM filter.}

\label{fig:main}
\vspace{-5mm}
\end{figure*}
\textbf{Visual Encoder.} Our video-based sign language translation framework is built around an image encoder that extracts spatial features \( F \) from input video frames \mbox{\( V = \{v_0, v_1, \dots, v_{T^{*}}\} \)}, where \( T^{*} \) represents the total number of frames. These features, with a dimensionality of \( D^{*} \), are structured as \mbox{$F_{video} \in \mathbb{R}^{T^{*} \times D^{*}}$}. We use and fine-tune DINOv2~\cite{dinov2learningrobustvisual} (ViT-S/14 variant), a self-supervised Vision Transformer, as our image encoder. The class token's output serves as a feature vector, which is transformed linearly and batch-normalized to obtain \mbox{\( F_{video} \)}. \\
\textbf{Feature Distillation.} In parallel to the visual encoder, sign videos are input to the HaMeR model to extract hand shape features. Specifically, we utilize the transformer head section of HaMeR, which was originally used to regress hand and camera parameters. In our work, we repurposed the transformer head to extract hand pose and global orientation parameters, as they were highly effective hand
shape descriptors for our task. For each frame, the hand pose features are represented as a tensor of \mbox{$H \in \mathbb{R}^{2\times15\times3\times3}$},  while the global orientation is parameterized as \mbox{$G \in \mathbb{R}^{2\times3\times3}$}. For simplicity and computational efficiency, we flatten these parameters, resulting in a 1D feature vector of \mbox{$H' \in \mathbb{R}^{270}$} for the hand pose and, \mbox{$G'\in \mathbb{R}^{18}$} for the global orientation. The flattened features are then fused to form \mbox{$F_{HaMeR} \in \mathbb{R}^{T^{*}\times288}$} for each input sign video, where \mbox{$T^{*}$} represents the total number of frames. 
To guide the visual encoder to extract hand features, we need to map \mbox{$F_{video}$} to \mbox{$F_{HaMeR}$} and then combine these mapped features to the original features in order not to lose other important features captured by the visual encoder:
\begin{align}
    F^{*}_{\text{video}} = \texttt{mapper}_{1}(F_{\text{video}}&) \in \mathbb{R}^{T^{*} \times 288}, \quad 
    F'_{\text{video}} = \texttt{mapper}_{2}(F^{*}_{\text{video}}) \in \mathbb{R}^{T^{*} \times D}. \label{eq:part1} \\
    \hat{F}_{\text{video}}& = \texttt{concat}(F'_{\text{video}}, F_{\text{video}}) \in \mathbb{R}^{T^{*} \times 2D}. \label{eq:part2}
\end{align}

For hand features distillation to the visual encoder, a simple $L_{2}$ loss can suffice. The distillation loss between $F^{*}_{\text{video}}$ and $F_{\text{HaMeR}}$ can be represented by:
\begin{equation}
    \mathcal{L}_{\text{distill}} = \frac{1}{T^{*} \times 288}\sum_{i}^{T^{*}} \sum_{j}^{288} \left\| F^{*}_{video,i,j} - F_{HaMeR,i,j} \right\|_2.
\end{equation}

where $T^{*}$ denotes the number of frames, and 288 is the HaMeR feature dimension. The feature \(\hat{F}_{\text{video}}\) is then passed into the temporal encoder for further processing.\\
\textbf{Temporal Encoder.} We use features \( \hat{F}_{video} \) as input to our temporal encoder to learn spatio-temporal sign representations. Since sign language sequences often contain hundreds of frames, we design a spatio-temporal transformer inspired by prior SLT approaches, incorporating two key modifications for efficiency and effectiveness. We integrate local self-attention with a window size of seven, a proven technique for SLT~\cite{gaslt}. Combined with downsampling, this extends the model’s temporal receptive field while reducing redundancy, ensuring compatibility with the LLM encoder. The resulting features are represented as \mbox{\( Z_{video} \in \mathbb{R}^{T \times D} \)}, where \( T \) is the downsampled sequence length and \( D \) is the new feature dimension. To effectively model both temporal and spatial information, we use Rotary Position Embedding (RoPE)~\cite{rope} for positional encoding.\\ 
\textbf{Video-Description Alignment.}
To focus on hand-centric temporal dynamics and distinguish signs more effectively, after the temporal encoder, we introduce a contrastive alignment between video and description representations. Rather than relying on a computationally expensive VideoLLM at inference time or discarding temporal information, we project the encoded video features into the description space and then combine them with the original video features to preserve both global and temporal context.
As illustrated in Fig.~\ref{fig:main}, each sign video is processed by a video description module, which generates a textual description. This description is then passed through a text encoder to obtain token-level features, $\hat{L} \in \mathbb{R}^{N\times K}$, where $N$ is the number of tokens and 
$K$ is the embedding dimension. Simultaneously, the video features $Z_{video}$ are mapped to the same feature space using a two-stage projection: 
\begin{align}
    Z^{*}_{\text{video}} = \texttt{mapper}_{3}(Z_{\text{video}}&) \in \mathbb{R}^{T \times D'}, \quad 
    Z'_{\text{video}} = \texttt{mapper}_{4}(Z^{*}_{\text{video}}) \in \mathbb{R}^{T \times D}. \label{eq:part1} \\
    \hat{Z}_{\text{video}}& = \texttt{concat}(Z'_{\text{video}}, Z_{\text{video}}) \in \mathbb{R}^{T \times D''}. \label{eq:part2}
\end{align}
The resulting feature \(\hat{Z}_{\text{video}}\) are forwarded to subsequent network components. To align video and description features for contrastive learning, we apply a linear projection if $D'' \neq k$, mapping both to a common dimension $\hat{D}$. We then use average pooling across the temporal and token dimensions to obtain global representations \mbox{$\tilde{Z}\in \mathbb{R}^{\hat{D}}$} for the video and \mbox{$\tilde{L}\in \mathbb{R}^{\hat{D}}$} for the description. The contrastive loss is computed as:
\begin{equation}
    \mathcal{L}_{\text{desc}} = -\frac{1}{2\times B}\Big(\sum^{B}_{j=1}log\frac{\exp(\texttt{sim}(\tilde{Z}_{j}, \tilde{L}_{j})/\tau_{1})}{\sum^{B}_{k=1}\exp(\texttt{sim}(\hat{Z}_{j}, \tilde{L}_{k})/\tau_{1})}
    + \sum^{B}_{j=1}log\frac{\exp(\texttt{sim}(\tilde{L}_{j}, \tilde{Z}_{j})/\tau_{1})}{\sum^{B}_{k=1}\exp(\texttt{sim}(\tilde{L}_{j}, \tilde{Z}_{k})/\tau_{1})}\Big).
\end{equation}
where \( \texttt{sim}(x, y) \) is the cosine similarity, and \( \tau_{1} \) is a learnable temperature parameter. \\
\textbf{Video-Target Alignment.}
To improve the effectiveness of converting video features into output text, we employ an encoder-decoder-based LLM. The previously constructed feature representation \mbox{$\hat{Z}_{video}$} is provided as input to the encoder, producing the transformed video features 
\mbox{$Y \in \mathbb{R}^{T \times K}$}, which are used for subsequent alignment.
To address the modality gap between video and language, we adopt a contrastive learning strategy inspired by CLIP~\cite{clip}. This approach aligns the video features with their corresponding spoken sentence representations, encouraging similarity between matched pairs while distinguishing them from unrelated ones.
Given a target sentence $TS$ associated with a video, we encode it into textual features \mbox{\( M \in \mathbb{R}^{\overline{N} \times K} \)} using a frozen text encoder, where \mbox{$\overline{N}$} is the number of tokens and $K$ is the embedding dimension. Average pooling is applied to both \mbox{$Y_{video} \in \mathbb{R}^{T \times K}$} and $M$ across temporal and token dimensions, respectively, resulting in global video and sentence features: 
\mbox{$\tilde{{Y}} \in \mathbb{R}^{K}$} and \mbox{$\tilde{M} \in \mathbb{R}^{K}$}.
We apply a symmetric contrastive loss as follows:
\begin{equation}
    \mathcal{L}_{\text{align}} = -\frac{1}{2\times B}\Big(\sum^{B}_{j=1}log\frac{\exp(\texttt{sim}(\tilde{Y}_{j}, \tilde{M}_{j})/\tau_{2})}{\sum^{B}_{k=1}\exp(\texttt{sim}(\hat{Y}_{j}, \tilde{M}_{k})/\tau_{2})}
    + \sum^{B}_{j=1}log\frac{\exp(\texttt{sim}(\tilde{M}_{j}, \tilde{Y}_{j})/\tau_{2})}{\sum^{B}_{k=1}\exp(\texttt{sim}(\tilde{M}_{j}, \tilde{Y}_{k})/\tau_{2})}\Big).
\end{equation}
where \( \texttt{sim}(x, y) \) is the cosine similarity, and \( \tau_{2} \) is a learnable temperature parameter. \\
The final pre-training loss is:  
\begin{equation}
\mathcal{L}_{\text{pre-train}} = \lambda_{\text{align}}\times\mathcal{L}_{\text{align}} + \lambda_{\text{desc}}\times\mathcal{L}_{\text{desc}} + \lambda_{\text{distill}}\times\mathcal{L}_{\text{distill}} 
\label{eq:loss}
\end{equation}
where \mbox{$\lambda_{\text{align}}$, $\lambda_{\text{desc}}$ and $\lambda_{\text{distill}}$} control the balance between these three losses. See the Appendix for a comprehensive evaluation conducted to determine the optimal weighting in the pre-training loss.\\
\textbf{Sign Language Translation.}
As illustrated in Fig.~\ref{fig:main}, 
the fine-tuning stage excludes the feature distillation, video-description alignment, and video-target alignment components. The mapping modules (purple), introduced in the feature distillation and video-description alignment components, learn to align feature spaces during pre-training. With the mapper modules, the feature distillation and video-description alignment components are no longer required after pre-training, eliminating the need for heavy frozen models such as the HaMeR transformer head, Video-LLM, and LLMs during fine-tuning. This reduction in computational overhead makes our framework more efficient and competitive with other approaches at inference time. During fine-tuning, the only newly introduced module is the LLM decoder, which is attached to the pre-trained LLM encoder and not initialized from pre-training.

Given a sign language video $V_i$, the decoder aims to generate the corresponding spoken sentence \mbox{$\widehat{TS}_i = (\widehat{TS}_{i,1}, \dots, \widehat{TS}_{i,T'})$}, where $T'$ denotes the number of tokens in the target spoken sentence. The decoder operates in an autoregressive manner; it begins with the special \texttt{<bos>} token and predicts tokens one by one until the \texttt{<eos>} token is reached.
The model is trained by minimizing the cross-entropy loss between the predicted tokens $\widehat{TS}_{i,j}$ and the ground truth tokens $TS_{i,j}$ as follows:
\begin{equation}
\mathcal{L}_{SLT_{i}} = - \sum_{j=1}^{T'} \log\hspace{2pt} p(\widehat{TS}_{i,j} \mid TS_{i,1:j-1}, V_{i}).
\end{equation}
\vspace{-1cm}
\section{Experiments}
\label{sec:results}
\subsection{Datasets and Evaluation Metrics}
\textbf{Datasets.} We conduct experiments on the Phoenix2014T~\cite{nslt} (or Phoenix14T) and CSL-Daily~\cite{csldaily} datasets for SLT, evaluating performance on their development and test sets. Phoenix-2014T is a German sign language dataset containing 2,887 unique German words, with 7,096 training samples, 519 validation samples, and 642 test samples. CSL-Daily is a Chinese sign language dataset with a vocabulary of 2,343 Chinese words, comprising 18,401 training samples, 1,077 validation samples, and 1,176 test samples.\\
\textbf{Evaluation Metrics.} We use BLEU~\cite{bleu} and ROUGE-L~\cite{rouge} to evaluate translation quality. BLEU-n measures n-gram precision~\cite{brown1992class}, and we report BLEU-1 to BLEU-4 scores. ROUGE-L (or ROUGE) evaluates the F1 score based on the longest common subsequence between predicted and reference translations. For simplicity, we denote BLEU-1 to BLEU-4 as B1, B2, B3, and B4, and ROUGE as R.
\vspace{-0.3cm}
\subsection{Implementation details}
\textbf{Model.} 
To address memory constraints, we apply LoRA~\cite{loralowrankadaptationlarge} adapters to the top three layers of the visual encoder. The adapters are configured with a rank of 4, a dropout rate of 0.1, scaling factor 4.0, and are applied to both self-attention and feed-forward weights.
The temporal encoder is a four-layer transformer with a hidden dimension of 512, 8 attention heads and an intermediate size of 2048, with temporal downsampling applied after the second layer. 
We adopt the \texttt{mBART-large-cc25} model~\cite{mbart}, which consists of 12 layers in both the encoder and decoder, as the backbone of our LLM-based encoder-decoder architecture. 
LoRA~\cite{loralowrankadaptationlarge} is applied to the LLM encoder and decoder modules, targeting $q_{\text{proj}}$ and $v_{\text{proj}}$ layers with a rank of 16, a scaling factor (LoRA alpha) of 32 and a dropout rate of 0.1. 
All mappers in Fig.~\ref{fig:main} are trainable and consist of a linear layer followed by \texttt{GELU} activation function.
Training is optimized with the XFormers~\cite{xFormers2022} library with Flash Attention~\cite{dao2023flashattention2fasterattentionbetter} for improved efficiency. In pre-training, we use \texttt{mBART-large-50}~\cite{mbart} as the text encoder for video-description alignment due to its broader multilingual coverage and robustness to noisy, diverse descriptions. For video-target alignment, we choose \texttt{mBART-large-cc25}~\cite{mbart} as the text encoder as it offers more stable embeddings for clean, well-structured target sentences in high-resource languages. We use \texttt{ShareGPT4Video-8B} ~\cite{chen2024sharegpt4videoimprovingvideounderstanding} as the video LLM in the pre-training stage to generate detailed and semantically rich hand-focused descriptions from sign language video segments, where each segment consists of 16 frames. We also use \texttt{GPT-4o-mini} API~\cite{openai2023gpt4} to generate a single coherent description, maintaining the temporal order of events and removing irrelevant or unnecessary information.\\
\textbf{Implementation.} The model is pre-trained for 100 epochs, and fine-tuned for 200 epochs, with a batch size of 16 on one A100 GPU. The input sequences are first resized into \mbox{$256 \times 256$}, and then randomly/centrally cropped into \mbox{$224 \times 224$} during training/inference.   
We use the \textit{AdamW}~\cite{adamw} optimizer with a learning rate of \mbox{$3 \times 10^{-4}$}
and weight decay of $0.001$. 

\begin{table*}[!t]
\tiny
\caption{\small Experimental results on Phoenix14T and CSL-Daily datasets. The best results for gloss-free models are highlighted in bold, while the second-best results are underlined. We abbreviate BLEU-1 to BLEU-4 as B1, B2, B3, and B4, and ROUGE as R.}
\centering
\resizebox{\linewidth}{!}{
\begin{tabular}{l|ccccc|ccccc}
\hline
\multirow{2}{*}{\textbf{Method}} &
\multicolumn{5}{c|}{\textbf{Phoenix14T}} &
\multicolumn{5}{c}{\textbf{CSL-Daily}} \\
\cline{2-11}
& \textbf{B-1} & \textbf{B-2} & \textbf{B-3} & \textbf{B-4} & \textbf{R}
& \textbf{B-1} & \textbf{B-2} & \textbf{B-3} & \textbf{B-4} & \textbf{R} \\
\hline
\rowcolor[HTML]{D2D2D2}
\multicolumn{11}{c}{\textbf{Gloss-based}} \\ \hline
MMTLB~\cite{mmtlb} & 53.97 & 41.75 & 33.84 & 28.39 & 52.65 & 53.31 & 40.41 & 30.87 & 23.92 & 53.25 \\
SLTUNET~\cite{sltunet} & 52.92 & 41.76 & 33.99 & 28.47 & 52.11 & 54.98 & 41.44 & 31.84 & 25.01 & 54.08 \\ 
TS-SLT~\cite{ts-slt} & 54.90 & 42.43 & 34.46 & 28.95 & 53.48 & 55.44 & 42.59 & 32.87 & 25.79 & 55.72  \\\hline
\rowcolor[HTML]{D2D2D2}
\multicolumn{11}{c}{\textbf{Gloss-free}} \\ \hline
NSLT+Luong~\cite{luong} & 29.86 & 17.52 & 11.96 & 9.00 & 30.70 & 34.16 & 19.57 & 11.84 & 7.56 & 34.54 \\
CSGCR~\cite{csgcr} & 36.71 & 25.40 & 18.86 & 15.18 & 38.85 & -- & -- & -- & -- & -- \\
GFSLT-VLP~\cite{gfslt-vlp} & 43.71 & 33.18 & 26.11 & 21.44 & 42.49 & 39.37 & 24.93 & 16.26 & 11.00 & 36.44 \\
Sign2GPT~\cite{sign2gpt} & 49.54 & 35.96 & 28.83 & 22.52 & 48.90 & 41.75 & 28.73 & 20.60 & 15.40 & 42.36 \\
SignCL~\cite{signcl} & 49.76 & 36.85 & \underline{29.97} & 22.74 & 49.07 & 47.47 & 32.53 & 22.62 & 16.16 & 48.92 \\
FLa-LLM~\cite{fla-llm} & 46.29 & 35.33 & 28.03 & 23.09 & 45.27 & 37.13 & 25.12 & 18.38 & 14.20 & 37.25 \\
SignLLM~\cite{signllms} & 45.21  & 34.78 & 28.05 & 23.40 & 44.49 & 39.55 & 28.13 & 20.07 & 15.75 & 39.91 \\
LLaVA-SLT~\cite{liang2024llavasltvisuallanguagetuning} & \underline{51.20} & \underline{37.51} & 29.39 & \underline{23.43} & \underline{50.44} & \underline{52.15} & \underline{36.24} & \underline{26.47} & \underline{20.42} & \underline{51.26} \\
\cdashline{1-11}
\textbf{BeyondGloss} & \textbf{52.38} & \textbf{38.57} & \textbf{30.74} & \textbf{25.49} & \textbf{52.89} & \textbf{53.12} & \textbf{38.63} & \textbf{27.82} & \textbf{21.53} & \textbf{53.46} \\
\hline
\end{tabular}}
\label{tab:results}
\end{table*}
\vspace{-5mm}
\subsection{Results}
\textbf{Results on Phoenix14T and CSL-Daily.} Tab. \ref{tab:results} compares gloss-based, and gloss-free methods on the Phoenix14T and CSL-Daily test datasets. Our method outperforms other gloss-free SLT approaches in terms of BLEU-4 and ROUGE.\\
\textbf{Qualitative Results.} Tab.~\ref{tab:quality} shows translation examples from Phoenix14T. We compare the reference translations with outputs from GFSLT-VLP~\cite{gfslt-vlp}, the only other publicly available gloss-free model, and our method. While GFSLT-VLP often captures only a few correct words or produces unrelated sentences, our method generates more accurate translations. These qualitative examples demonstrate the effectiveness of our approach.

\begin{table*}[t]
\tiny
\caption{\small Qualitative results of Phoenix14T. \colorbox{green!30}{Green} means totally same as the reference.}
\centering
\begin{tabular}{l p{0.6\linewidth}}
\hline
\textbf{Reference:} & am tag wechseln sonne und wolken einander ab teilweise ist es auch langere zeit sonnig . \\
& (During the day sun and clouds alternate partly it is sunny for a long time) \\[1ex]
\textbf{GFSLT-VLP:} & \colorbox{green!30}{am tag wechseln sonne und wolken einander ab} es bilden sich \colorbox{green!30}{langere zeit} viel sonnenschein. \\ & (During the day, sun and clouds alternate There is a lot of sunshine for a long time)  \\[1ex]
\textbf{BeyondGloss:} & \colorbox{green!30}{am tag wechseln sonne und wolken einander ab} zeigt sich die \colorbox{green!30}{auch langere zeit sonnig .} \\ & (During the day, sun and clouds alternate, and the weather remains sunny for longer periods.) \\ \hline
\textbf{Reference:} & am tag nur hier und da einige sonnige momente vor allem an den alpen . \\ & (During the day only here and there some sunny moments, especially in the Alps)\\[1ex]
\textbf{GFSLT-VLP:} & morgen zeigt sich mal die sonne wenn dann \colorbox{green!30}{vor allem an den alpen.} \\ & (Tomorrow the sun will show up , especially in the Alps) \\[1ex]
\textbf{BeyondGloss:} & und morgen scheint \colorbox{green!30}{und da einige} die sonnige \colorbox{green!30}{vor allem an den alpen .} \\ & (Tomorrow will shine and there will be some sunny weather, especially in the Alps .) \\ \hline
\textbf{Reference:} & am freitag scheint abseits der nebelgebiete haufig die sonne . \\ & (On Friday the sun often shines outside of the foggy areas)\\[1ex]
\textbf{GFSLT-VLP:} & \colorbox{green!30}{am freitag} sobald der nebel weg ist sonnenschein. \\ & (On Friday as soon as the fog is gone there will be sunshine)  \\[1ex]
\textbf{BeyondGloss:} & \colorbox{green!30}{am freitag scheint abseits der nebelgebiete haufig die sonne .} \\ & (On Friday the sun often shines outside of the foggy areas) \\ \hline
\end{tabular}
\label{tab:quality}
\vspace{-5mm}
\end{table*}
\vspace{-5pt}
\subsection{Ablation studies}
\textbf{Analysis of Pre-Training Components Contributions.} We conduct experiments to evaluate the impact of the core components of our proposed method. Tab.~\ref{tab:comp} reports the performance obtained by progressively incorporating each component. As shown in the first row, omitting pre-training and alignment between sign videos and output texts results in significantly lower downstream performance, highlighting the importance of pre-training for SLT. The second row demonstrates that video-target alignment contributes substantially to performance improvement. Rows three and four further show the positive effect of our proposed video-description alignment. In rows four and five, feature distillation also contributes to modest improvements in performance. Finally, combining all components together (row five) yields the best overall performance, achieving state-of-the-art results.\\
\textbf{Evaluating Different Visual Encoders.} We conduct an additional ablation study to examine the impact of different visual encoders on final performance. As shown in Tab.~\ref{tab:visual}, using DINOv2 \cite{dinov2learningrobustvisual}, a transformer-based visual encoder, yields better results compared to ResNet-18 \cite{he2016deep}, a convolutional encoder. 
The best performance is achieved when applying LoRA to the last three layers of DINOv2, leading to our state-of-the-art results. This is because DINOv2 has not been trained on sign language data, and fine-tuning the last three layers allows it to adapt to our domain while only slightly modifying its weights.\\
\textbf{Evaluating Different LLMs.} As shown in Tab.~\ref{tab:llm}, we assess the impact of using various LLMs within our framework. Both MT5-Base~\cite{xue2021mt5} and MBART (24-layer)~\cite{mbart}, which have a comparable number of parameters, yield strong results. MBART slightly outperforms MT5-Base, likely due to its stronger multilingual prior knowledge. These findings highlight the generalisation ability of our approach. Additionally, we evaluate a 6-layer version of MBART, similar to the configuration used in GFSLT-VLP~\cite{gfslt-vlp}, which achieves moderate performance.
It is worth mentioning that all the tests in the ablation studies are conducted on the Phoenix14T validation set.
\begin{table*}[]
\caption{\small Ablation study on key elements in our framework.}
\centering
\resizebox{0.6\linewidth}{!}{
\begin{tabular}{c|ccc|cc}
\hline
 & \makecell{\textbf{Feature} \\ \textbf{Distillation}} & \makecell{\textbf{Video-Description}\\\textbf{Alignment}} & \makecell{\textbf{Video-Target}\\ \textbf{Alignment}} & \textbf{B-4} & \textbf{R} \\ \hline
 (1) & \textbf{--}& \textbf{--}& \textbf{--}& 16.31 & 36.33\\
 (2) & \textbf{--}& \textbf{--}& \checkmark & 21.87  & 44.30 \\
 (3) & \textbf{--}& \checkmark & \checkmark & 23.47 & 47.17 \\
 (4) & \checkmark& \checkmark & \textbf{--} & 22.47 & 44.87 \\
  (5) & \checkmark   &   \checkmark   &  \checkmark    & \textbf{25.68} &
  \textbf{53.85}   \\
  \hline
\end{tabular}}
\label{tab:comp}
\end{table*}

\begin{table*}[!t]
\centering
\begin{minipage}{0.48\textwidth}
\centering
\caption{\small Ablation study on visual encoder.}
\resizebox{\linewidth}{!}{
\begin{tabular}{cc|cc}
\hline
\textbf{Visual Encoder} & \textbf{Trainable} & \textbf{B-4} & \textbf{R} \\ \hline
ResNet-18\cite{he2016deep}& \checkmark  & 18.97 & 40.17   \\
DINOv2\cite{dinov2learningrobustvisual}& $\times$ & 20.43 & 42.75   \\
DINOv2\cite{dinov2learningrobustvisual}& 3 last layers fine-tuned by LoRA & \textbf{25.68} & \textbf{53.85}\\
\hline
\end{tabular}}
\label{tab:visual}
\end{minipage}
\hfill
\begin{minipage}{0.48\textwidth}
\centering
\caption{\small Ablation study on LLM.}
\resizebox{\linewidth}{!}{
\begin{tabular}{cc|cc}
\hline
\textbf{Model} & \textbf{Num of Layers/Num of parameters} & \textbf{B-4} & \textbf{R} \\ \hline
  MBart~\cite{mbart} & 6 / \textasciitilde115M  & 22.50  & 45.70  \\
  MT5-Base~\cite{xue2021mt5}   & 24 / \textasciitilde580M & 24.18   & 49.87 \\
  MBart~\cite{mbart} & 24 / \textasciitilde600M & \textbf{25.68} & \textbf{53.85} \\
  \hline
\end{tabular}}
\label{tab:llm}
\end{minipage}
\end{table*}
\vspace{-4mm}
\section{Conclusion}
\label{sec:conclude}
In this work, we introduced \textbf{BeyondGloss}, a novel gloss-free Sign Language Translation framework that emphasizes hand-centric modeling and leverages the capabilities of VideoLLMs. By generating fine-grained, temporally-aware textual descriptions of hand motion and aligning them with sign video features through contrastive learning, our approach effectively bridges the modality gap between visual and linguistic representations. Additionally, the integration of detailed hand features distilled from HaMeR further enriches the sign video representations. Extensive experiments on Phoenix14T and CSL-Daily demonstrate that \textbf{BeyondGloss} achieves state-of-the-art performance, underscoring the importance of structured hand modeling and multimodal alignment in advancing gloss-free SLT.
\section*{Acknowledgement}
    This work was supported by the SNSF project ‘SMILE II’ (CRSII5 193686), the Innosuisse IICT Flagship (PFFS-21-47), EPSRC grant APP24554 (SignGPT-EP/Z535370/1), and through funding from Google.org via the AI for Global Goals scheme. This work reflects only the author’s views, and the funders are not responsible for any use that may be made of the information it contains.

\bibliography{egbib}
\clearpage
\section*{Supplementary Material}
\section*{Sign Video Descriptor}

\subsection*{Video Segmentor Algorithm}
\begin{algorithm}
\caption{Video Segmentation into 16-Frame Segments}
\label{alg:video_segmentor}
\begin{algorithmic}[1]
\REQUIRE Video file $V$
\ENSURE List of 16-frame segments $S$

\STATE Open video $V$ and get total frames $N$
\STATE Initialize empty list $S \gets [\,]$
\STATE $segment \gets [\,]$ \COMMENT{Current segment buffer}

\FOR{$i \gets 1$ to $N$}
    \STATE $frame \gets$ read next frame from $V$
    \STATE Append $frame$ to $segment$
    
    \IF{$\text{length}(segment) == 16$}
        \STATE Append $segment$ to $S$
        \STATE Reset $segment \gets [\,]$ \COMMENT{Start new segment}
    \ENDIF
\ENDFOR

\IF{$\text{length}(segment) > 0$}
    \STATE Append remaining $segment$ to $S$ \COMMENT{Handle leftover frames}
\ENDIF

\RETURN $S$
\end{algorithmic}
\end{algorithm}

\subsection*{More Examples}
\label{append:more}
\begin{figure*}[!t]
\begin{center}
\includegraphics[width=\linewidth]{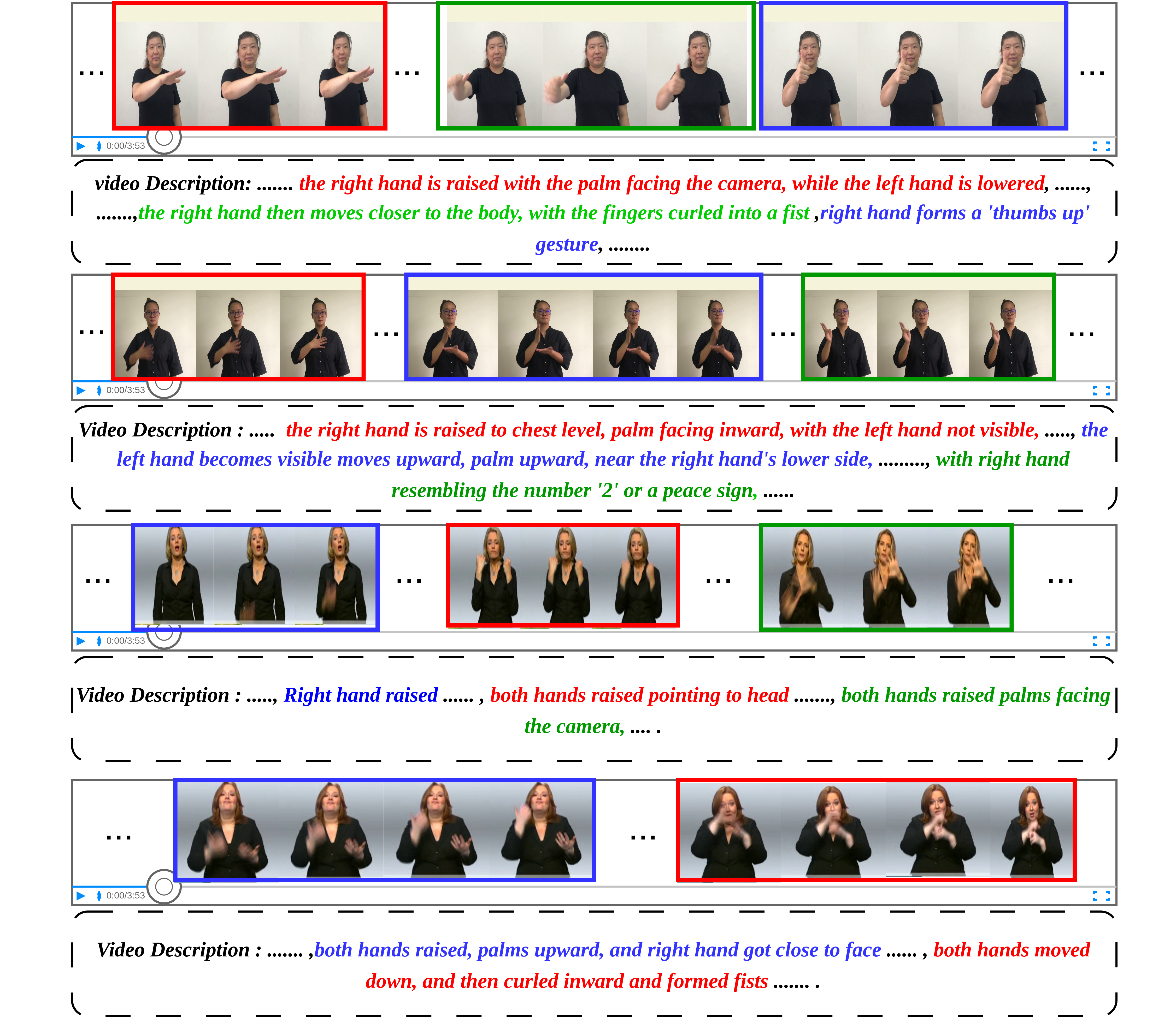}
\end{center}
   \caption{More examples of video description generated by the proposed Sign Video Descriptor in Phoenix-2014T and CSL-Daily. The rectangle around the sequence of frames and its corresponding description in the final video description share the same color.}
\label{fig:more-examples}
\end{figure*}

We show more examples of video description generated by the proposed Sign Video Descriptor in Fig.~\ref{fig:more-examples}. 
\section*{Ablation studies}
\subsection*{Impact of Loss Weighting in Pre-training for SLT}
As shown in Tab.~\ref{tab:abblate_comp}, we have conducted a comprehensive evaluation to determine the optimal weighting in the pre-training loss.
\begin{table}[H]
\small
\caption{Impact of Loss Weighting in Pre-training for SLT.}
\centering
\resizebox{0.6\linewidth}{!}{
\begin{tabular}{c|ccc|cc}
\hline
 & $\lambda_{\text{distill}}$ & $\lambda_{\text{desc}}$ & $\lambda_{\text{align}}$ & \textbf{B-4} & \textbf{R} \\ \hline
 (1) & 0.0 & 0.0 & 0.0 & 16.41 & 36.33\\
 (2) & 0.0 & 0.0 & 1.0 & 21.87  & 44.30 \\
 (3) & 0.0 & 1.0 & 1.0 & 22.13 & 45.14 \\
 (4) & 0.0 & 0.5 & 1.0 & 23.47 & 47.17 \\
 (5) & 1.0 & 1.0 & 1.0 & 22.85 & 46.32 \\
 (6) & 0.5 & 1.0 & 1.0 & 24.56  & 50.21 \\
 (7) & 0.3 & 0.5 & 1.0 & \textbf{25.68} & \textbf{53.85} \\ 
  \hline
\end{tabular}}
\label{tab:abblate_comp}
\end{table}

\end{document}